\documentclass[11pt,a4paper]{article}
\usepackage[margin=2.5cm]{geometry}
\usepackage{amsmath,amssymb}
\usepackage{graphicx}
\usepackage{xcolor}
\usepackage[colorlinks=true,linkcolor=blue!55!black,urlcolor=blue!55!black,citecolor=blue!55!black]{hyperref}
\usepackage{array}
\usepackage{multirow}
\usepackage{booktabs}
\usepackage{enumitem}
\usepackage{float}
\usepackage{caption}
\usepackage{authblk}
\usepackage{natbib}
\usepackage{tabularx}

\newcommand{\vx}{\mathbf{x}}

\newcommand{\vphi}{\boldsymbol{\phi}}

\title{Shared Symbolic Backbones for Physically Consistent Multi-Output Symbolic Regression}
\author{Manuel Rodriguez}
\affil{Chemical and Environmental Engineering Department. Universidad Politecnica de Madrid}
\date{July 12, 2026}

\begin{document}
\maketitle

\begin{abstract}
Symbolic regression provides analytical expressions, but it is usually applied one output at a time. This is limiting in process systems, where state variables are often coupled through shared physical parameters. Independent symbolic regression can give accurate individual equations that are difficult to interpret as one model. We present a neuro-evolutionary symbolic-regression method for coupled multi-output systems. The method searches for a shared symbolic backbone: a set of latent symbolic units that is discovered once and reused by several outputs through sparse additive or multiplicative read-outs. The discrete model structure is evolved by mutation and crossover, whereas the continuous parameters are tuned by gradient descent and inherited by the offspring.

The method is assessed on a set of benchmarks with known ground truth and on a hydrothermal-liquefaction yield case. The results show that coupling is not a general route to lower prediction error. Its main contribution is the enforcement and diagnosis of cross-output consistency when a physically shared factor is embedded in a latent expression and is weakly identifiable from the data. This occurs for Langmuir--Hinshelwood and site-coverage denominators, for which independent PySR does not close the consistency gap or recover the same shared form. Conversely, when each output is already identifiable, as in the Van de Vusse benchmark, independent symbolic regression matches or improves the coupled model. The proposed framework, rather than a general-purpose predictor, is a structured shared-mechanism extractor. Its value is highest when the target structure is sparse, shared, weakly identifiable or constrained by closure.
\end{abstract}

\section{Introduction}

Process systems rarely produce independent outputs. Reaction rates may contain the same Arrhenius factor, adsorption denominator or inhibition term; product yields are linked by mass closure; dynamic state variables are coupled through stoichiometry and conservation laws. Yet data-driven symbolic regression (SR) is almost always applied one output at a time. Fitting each output separately makes the regression simpler, but it can produce sets of equations that are accurate individually and incoherent when read together: two rates that should share one temperature dependence may be assigned different apparent activation energies, and separately fitted yield correlations may reproduce each fraction yet violate closure when evaluated jointly. The issue is therefore not only prediction error. A set of independent symbolic models can each have small error while failing to constitute a single, physically consistent process model. What is often wanted in an engineering context is the opposite of a black-box fit: a compact set of equations in which a shared physical factor is represented once, consistently, and can be read, simplified and checked against mechanistic knowledge. This motivates searching for equations at the level of the coupled system rather than the individual output.

This paper formulates and tests a shared-backbone symbolic-regression method for coupled multi-output systems (Multi-Output Shared-Backbone Neuro-Evolutionary Symbolic Regression, MO-SB-NESR). The method searches for a finite set of reusable symbolic latent expressions; each output selects a sparse subset of these and combines them through a multiplicative read-out (suitable for rate laws) or an additive read-out (suitable for balances and yield correlations). The discrete symbolic structure is evolved, and the continuous parameters are tuned by gradient descent and inherited by the offspring.

We deliberately do not claim that multi-output coupling is generally more accurate than independent SR; the experiments show that it is not. The relevant question is narrower and, for process applications, more useful: \emph{under which conditions does a shared backbone enforce a cross-output physical consistency that independent symbolic regression cannot?} The answer that emerges from the benchmarks is specific. The advantage appears when a shared parameter is embedded inside a shared latent expression and is weakly identifiable from the data of each output alone. When each output identifies the same parameter on its own, coupling is unnecessary and independent SR matches or improves on it.

\paragraph{Contributions.}
\begin{enumerate}[leftmargin=*]
\item A shared symbolic-backbone architecture for coupled multi-output symbolic regression, with additive and multiplicative read-outs and literal, inspectable sharing of latent units.
\item A neuro-evolutionary training procedure in which evolution searches the discrete structure (operators, input masks, read-out connections, modes) and gradient descent tunes the continuous parameters under Lamarckian inheritance.
\item A structured-sparsity mechanism that makes selective sharing emerge, avoiding the uniform entanglement produced by rewarding sharing directly.
\item A three-arm prior protocol (unconstrained, soft-seeded, hard-frozen) that separates blind discovery from reachability-after-seeding and from prior-imposed confirmation, and that audits what injected domain knowledge actually contributed.
\item A benchmark-based characterization of when the shared backbone helps (per-output non-identifiable shared structure) and when independent SR is sufficient, including the sharpening of the criterion from ``non-separable'' to ``per-output non-identifiable.''
\item A grouped-data application (hydrothermal-liquefaction yields) demonstrating the method as a closure-respecting shared-structure extractor, with an explicit statement of its predictive limits.
\end{enumerate}

The manuscript is organized as follows. Section~\ref{sec:related} reviews the work with respect to symbolic regression and physics-aware neuro-evolution. Section~\ref{sec:method} describes the architecture, training procedure, and evaluation. Section~\ref{sec:benchmarks} presents the different benchmarks and the results, proceeding case by case so that each result is interpreted in relation to the type of problem tested. Section~\ref{sec:htl} applies the method to a hydrothermal-liquefaction yield dataset and Section~\ref{sec:conclusions} concludes.

\section{Related work}
\label{sec:related}

\subsection{Symbolic regression and its multi-output gap}

Symbolic regression searches for explicit mathematical expressions that optimise a fit criterion over data, without prescribing the functional form in advance. Classical implementations use genetic programming (GP), where candidate expression trees are evolved by mutation and crossover operators. The use of symbolic regression in chemical and process engineering allows one to have an equation that can be inspected, simplified, and checked against domain knowledge instead of having black-box regressors that are not interpretable.

A widely used open-source SR tool is PySR \citep{cranmer2023}, which combines multi-population evolutionary search with simplification rules to discover scalar expressions efficiently. More recent neural approaches attempt to learn the entire analytical model structure and coefficients using regularised gradient descent. The Equation Learner (EQL) \citep{martius2016,sahoo2018} defines a differentiable feed-forward network whose activation functions are algebraic primitives, enabling differentiation through the expression structure. Kolmogorov--Arnold networks (KAN) have recently been used for symbolic regression \citep{panczyk2025} applied to the energy industry. Class Symbolic Regression \citep{tenachi2024} combines deep reinforcement learning with in situ dimensional analysis constraints to automatically find a single analytical functional form that accurately fits multiple datasets. Other approaches use transformers like \citet{kamienny2022}, \citet{otte2026}. In Lu's approach, Deep Differentiable Symbolic Regression Neural Network (DDSR-NN) every layer represents a continuous distribution of mathematical operators, and the connections between layers represent the coefficient between these operators \citep{lu2025}. However, all of these methods remain fundamentally scalar-output tools: multi-output problems are handled by repeating the regression independently for each target.

An alternative paradigm draws on sparse identification of nonlinear dynamics (SINDy) \citep{brunton2016} and library-based regression, where a fixed dictionary of candidate terms is used and sparsity-promoting regression selects the active terms. SINDy is naturally extendable to multi-output systems through simultaneous coefficient identification, but it requires a pre-specified library and cannot discover novel functional forms such as adsorption denominators or Arrhenius factors. Some approaches are tailored to specific problems, like ADoK (Automated Discovery of Kinetic rate models) and its physics-informed extension PI-ADoK for catalytic kinetics discovery \citep{decarvalho2024,decarvalho2025}, demonstrating that constraint-guided search substantially reduces the experimental burden for kinetic model convergence. These works confirm both the opportunity and the difficulty of automated kinetic model discovery, but they remain single-output in their core formulation.

These methods are powerful, but with few exceptions they are formulated for scalar outputs. A multi-output problem is then handled by running the method once per output. For process systems this independent treatment is unsatisfactory in a specific way: the same physical mechanism recurs across equations --- a rate contributes to several balances, several catalytic rates share one adsorption denominator, product fractions obey a closure relation --- and independent fitting has no mechanism to represent that recurrence once and consistently. The result is a set of expressions that may each be accurate yet collectively encode the same factor in different, mutually incompatible forms.

\subsection{Neuro-evolution and physics-aware search}

The specific neuro-evolutionary paradigm adopted here was introduced for physics-aware SR by \citet{kubalik2025}, who proposed combining evolutionary search over neural network topologies with gradient-based tuning of network parameters. Their EN4SR method demonstrated that evolution over discrete structure escapes the entanglement failures of gradient-only approaches, which reach high values while leaving latent representations as linear mixtures of true factors. This separation of concerns --- evolution for structure, gradient descent for parameters --- is the substrate on which the shared-backbone architecture is built.

The specific step taken here is to make the symbolic representation shared across outputs. The broader machine-learning literature on multi-task learning (MTL) establishes that shared representations improve generalisation when tasks are related through a common latent structure \citep{lachapelle2023} and that disentanglement is not identifiable without inductive bias \citep{locatello2019}. Rather than evolving one expression per output, the method evolves a finite backbone of symbolic latent units together with sparse, output-specific read-outs. This connects to a result from multi-task representation learning: sparse task-specific predictors over a common representation can induce a disentangled representation, where the different outputs play the role of tasks and the backbone plays the role of the shared representation. Sharing is therefore not rewarded directly --- which we find collapses to uniform entanglement --- but is allowed to emerge from per-output read-out sparsity. In the machine-learning context, semantics-guided multi-task genetic programming \citep{wang2025} is the closest existing work to the method proposed here. That approach also applies genetic programming to multi-output regression with shared semantic building blocks, but focuses on predictive accuracy rather than on enforcing physical consistency of a shared mechanistic factor, and does not handle the weak-identifiability regime characterised in this paper.

The present work extends this paradigm specifically to make the symbolic representation shared across outputs. Rather than evolving one expression per output, the method evolves a finite backbone of symbolic latent units together with sparse output-specific read-out masks. This is qualitatively different from running neuro-evolutionary SR independently for each output: the backbone is evolved once, and cross-output evidence is pooled when evaluating any individual structure.

\section{Shared-backbone methodology}
\label{sec:method}

\subsection{Architecture: the shared symbolic backbone}

\begin{figure}[H]
\centering
\includegraphics[width=0.95\textwidth]{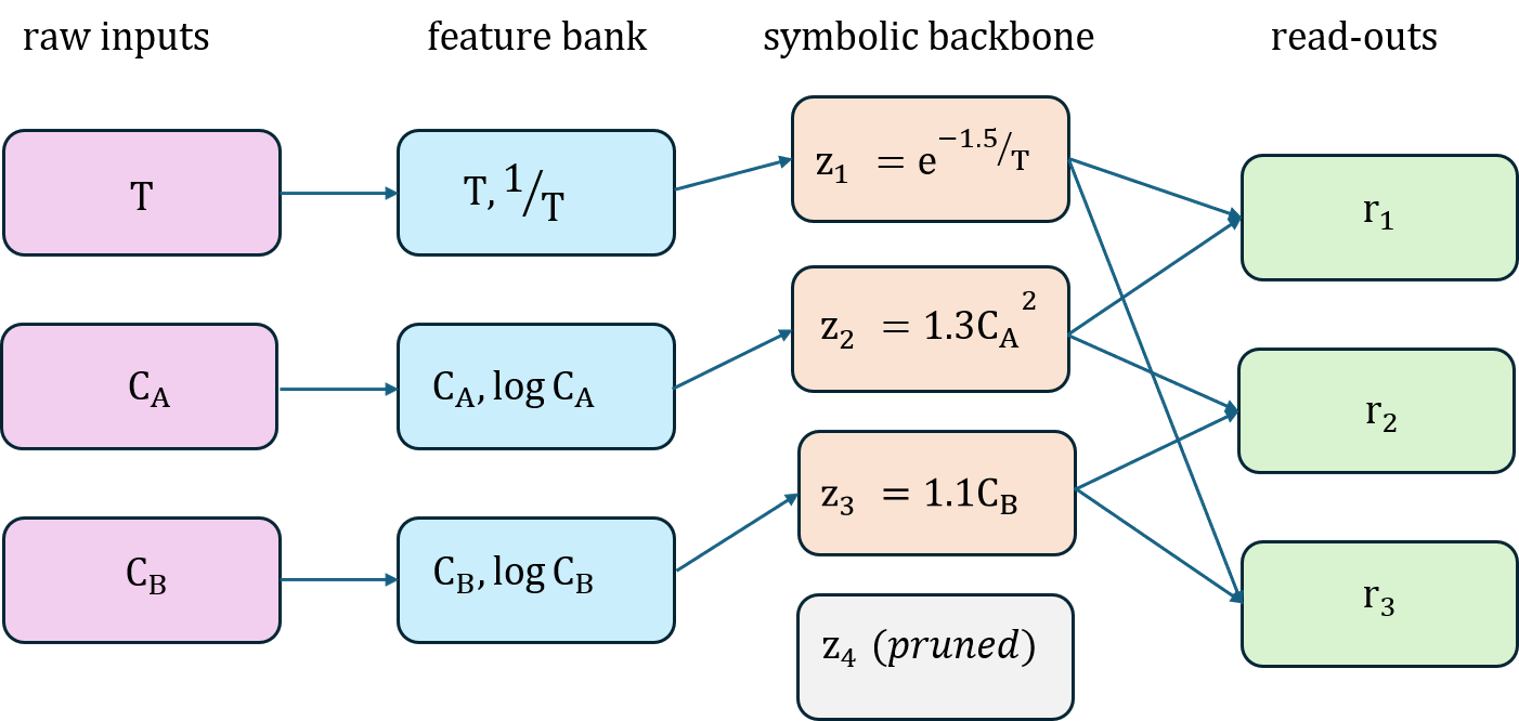}
\caption{The method architecture. Raw inputs are transformed into chemistry-friendly features, combined into a shared row of reusable building-block ``latent units'', the symbolic backbone, from which each output selects a few and combines them into its final equation. The backbone is shared: a factor discovered once is reused across outputs.}
\label{fig:architecture}
\end{figure}

The architecture is formalized through three components. First, the feature bank creates transformed inputs. Second, the symbolic backbone where a set of latent units define reusable symbolic pieces. Third, the read-out mask determines which outputs use each latent, and the additive or multiplicative read-out converts the selected latents into final equations. Figure~\ref{fig:architecture} shows the proposed architecture.

\paragraph{Feature bank.} Inputs are expanded into a configurable primitive feature bank
\begin{equation}
\vphi(\vx) = [\,x,\ 1/x,\ \log x,\ \ldots\,], \qquad \vphi \in \mathbb{R}^{n_f},
\end{equation}
with each block optional.

\paragraph{Symbolic backbone.} The backbone is a single layer of $H$ symbolic units, latents. Unit $h$ has a discrete operation $\sigma_h \in \mathcal{O}$, a binary input mask $s_h \in \{0,1\}^{n_f}$, and continuous parameters $(w_h, b_h)$:
\begin{equation}
z_h = \sigma_h\!\left(w_h^{\top}(s_h \odot \vphi(\vx)) + b_h\right), \qquad
\mathcal{O} = \{\exp,\ (\cdot)^2,\ \mathrm{id},\ 1/(\cdot),\ \log(1+\cdot),\ \sin,\ \sqrt{\cdot}\,\}.
\end{equation}
A single latent combines some features and wraps the result in one operation chosen from a set of operations. The latents are shared by all the outputs that need them: there is one copy of each latent, not a different one per output.

\paragraph{Read-outs.} Each output $o$ carries a binary mode and a read-out mask row $M_{o,\cdot} \in \{0,1\}^{H}$. The output builds its final equation by selecting a few latents and combining them in one of two ways: multiplicatively (as in rate laws) or additively (as in balances).
\begin{align}
\text{multiplicative (power-law rate form):} \quad & y_o = k_o \prod_{h=1}^{H} z_h^{\,e_{oh} M_{oh}}, \\
\text{additive (species-balance form):} \quad & y_o = \kappa_o + \sum_{h=1}^{H} c_{oh}\, M_{oh}\, z_h,
\end{align}
with learnable exponents $e_{oh}$ and coefficients $c_{oh}$. Sharing is literal: a latent multiplied (or summed) into several outputs is a shared factor, with output-specific exponents carrying parametric differences --- e.g. one Arrhenius latent $e^{-1/T}$ with exponents $(1, 2.5, 1)$ represents three activation energies from one discovered form. The model \emph{is} the formula: expressions are printed directly from the genome and weights. That constitutes the model. The final answer obtained is these equations, there is no hidden network to trust.

\subsection{Neuro-evolutionary training}

The central question: how does a slot ``decide'' to be an exponential rather than a square? It is done by an evolutionary search where a population of $P$ individuals is evolved for $G$ generations.

The procedure is the following. \textbf{(1) Start a population} --- not one model but dozens, each with a random backbone. \textbf{(2) Score each one}. \textbf{(3) Selection}: keep the better-scoring candidates, discard the worse. \textbf{(4) Breed} the survivors two ways: \emph{mutation} makes a small random change to one candidate (flip a latent from square to exponential, change which inputs feed a slot, switch an output between multiplicative and additive); \emph{crossover} swaps whole latents between two good candidates, so a useful Arrhenius factor found in one lineage can spread into another. \textbf{(5) Repeat} for hundreds of generations. The population drifts toward backbones whose operations match the physics in the data, because candidates that placed the right operation in the right slot survived and reproduced.

The genetic algorithm takes care of the structure and the neuro part takes care of the numerical values. Once a candidate's structure is fixed, finding its best numbers is a smooth optimization, gradient descent is used, so the model once it has a structure fixed behaves like a neural network whose weights are the physical constants. So neuro-evolutionary means evolution searches the structure and neural-network-style training tunes the constants.

The genotype used by the genetic algorithm (discrete part) is Unit ops $\{\sigma_h\}$, input masks $\{s_h\}$, read-out mask $M$, and per-output mode bits. There is where mutation flips read-out edges and input-mask bits and crossover swaps whole latent units between parents. The phenotype are the parameters (continuous part) that are tuned by Adam optimization with the following normalized per-output loss:
\begin{equation}
\mathcal{L} = \frac{1}{n\, n_{\mathrm{out}}} \sum_{i,o} \frac{\left(\hat{y}_o(x_i) - y_{io}\right)^2}{\sigma_o^2}, \qquad
\sigma_o^2 = \mathrm{Var}_i\, y_{io}.
\end{equation}
When a good candidate breeds, its children inherit the tuned numbers, not random ones. If a parent has tuned an Arrhenius factor to $\exp(-1.5/T)$, its offspring start from $-1.5$. This is like using the best conditions from one experimental campaign as the starting point for the next, rather than restarting blind each time.

\subsection{How models are evaluated}

There is no single ``classic'' score, because two different parts are considered to evaluate the fitness of the model.

During the search each candidate receives a fitness with two added parts. The first is how well it fits, measured per output relative to each output's own variability, the second is how simple it is: penalties for using too many latents, too many connections, too many inputs per latent. This simplicity pressure is what makes sharing selective because an output will not select a latent unless the fit improvement is worth the penalty, so latents are shared only when they are really needed. The following expression is used to evaluate the fitness:
\begin{equation}
F = \log \mathcal{L} + \lambda_{\mathrm{row}} \sum_{o,h} M_{oh}
+ \lambda_{\mathrm{lat}} \left|\{h : \exists o,\ M_{oh} = 1\}\right|
+ \lambda_{\mathrm{in}} \sum_{h\ \mathrm{active}} \|s_h\|_1.
\end{equation}
where $\log \mathcal{L}$ is the commented loss, $\lambda_{\mathrm{row}} \sum_{o,h} M_{oh}$ is the term pushing for sparse per-output reads, it is the term that drives toward disentanglement, having less connections, $\lambda_{\mathrm{lat}} |\{h : \exists o,\ M_{oh} = 1\}|$ penalizes how many distinct latent units the model uses at all, and $\lambda_{\mathrm{in}} \sum_{h\ \mathrm{active}} \|s_h\|_1$ penalizes latents that depend on many inputs, pushing each latent toward a simple, few-input sub-expression. The three penalties together target three independent axes of complexity --- how many connections (Term 2), how many latents (Term 3), how complex each latent (Term 4) and the loss is balanced against these three complexity penalties.

For choosing the final model, non-fitness criteria are also taken into account. The fitness score cannot fairly compare models built under different settings. The final choice implemented uses two penalty-free quantities: error on held-out validation data (data not used in fitting, so it measures genuine generalization rather than memorization) and a plain count of complexity (how many pieces the equation has). Among all candidates the search produced, one plots error against complexity and picks the simplest model within a small tolerance of the best error, avoiding complexity that is not needed.

So there is no classic evaluation, a model is judged on fit and simplicity and generalization to held-out data and, where the true answer is known, on whether it recovered the right structure.

\subsection{Optional domain priors and the three-arm protocol}
\label{sec:threearm}

The method runs without any prior knowledge, but mechanistic knowledge, when available, can be injected. The following protocol is devised to establish that the model produced was discovered from the data and not handed to the method.

Domain knowledge enters as an optional set of prior units that are symbolic units with a pinned operation and input set. With no prior, the search is unchanged: the empty prior reproduces unconstrained search exactly. A prior unit can be supplied in two different ways:
\begin{itemize}[leftmargin=*]
\item \textbf{soft} --- the unit is seeded into the initial population, but evolution is free to keep or discard it. This asks: if the structure is offered, can the search reach and retain it?
\item \textbf{hard} --- the unit's operation and inputs are frozen, and only its constants are fitted. This asks only: given the mechanism as fact, does it fit?
\end{itemize}

The protocol runs all three arms together. The reason is that a hard-arm success alone is nearly uninformative. Freezing a structure and fitting its constants proves only that the assumed form can fit; it cannot distinguish a real mechanism from one imposed by the analyst. The three arms are run together reaching:
\begin{itemize}[leftmargin=*]
\item \textbf{all-agree}: unconstrained, soft and hard all recover the form, the mechanism is supported by the data alone, no prior needed.
\item \textbf{soft-recovers}: soft matches hard, but unconstrained does not, the structure is reachable once seeded but not found blind. A reachability gap that is closed with the hint provided.
\item \textbf{hard-only}: only the frozen arm shows the form --- the data cannot support it even when offered as a seed. An identifiability limit; better data, not more search, is required.
\item \textbf{disagree}: the arms settle on incompatible forms at similar error --- the prior is fighting the data.
\end{itemize}

The soft arm is what makes this diagnostic possible: it separates reachability (the search misses a basin it could occupy) from identifiability (the data cannot support the form at all), a distinction that unconstrained-versus-hard alone cannot draw, because the hard arm forces the answer regardless.

\section{Benchmark testing}
\label{sec:benchmarks}

This section describes the different benchmarks used and the criteria to assess them. The benchmarks are synthetic, so they have known ground truth. All the expressions and shared parameters are known and are used to evaluate not only prediction error, but also whether the recovered expressions contain the correct shared factors, whether shared parameters are estimated consistently across outputs, and whether the final equations remain compact enough to be inspected, which is not possible with real experimental data alone. Unless otherwise indicated, datasets contain approximately 1500 samples split into 1200 training and 300 validation points.

\subsection{Protocol, baselines and evaluation criteria}

The coupled model is compared with three reference levels. The first is a log-linear least-squares baseline, used only for benchmarks whose true expressions admit such a transformation. This baseline is not a symbolic-regression competitor in the general case, but it is useful for identifying cases that are already easy by classical methods. The second is an uncoupled version of the same neuro-evolutionary machinery, with one private backbone per output; this isolates the effect of sharing. The third is independent PySR, run separately for each output, which represents a strong state-of-the-art symbolic-regression baseline.

Four criteria are used throughout the section. Prediction quality is measured by validation MSE or per-output normalized validation MSE when output scales differ. Complexity $C$ counts active read-out edges, active latent units and active input links. Structural recovery is assessed by comparing recovered latents with known primitives up to permutation and monotone transformations. Cross-output consistency is measured by the disagreement between estimates of the same physical parameter (for example in kinetic benchmarks the activation energies).

Evolutionary symbolic regression is seed-sensitive. For this reason, selected models are accompanied, where possible, by success rates over multiple restarts. A run is counted as successful only when it recovers the intended form and passes a Case-1 cleanliness check based on projection onto the true basis. The purpose is to avoid presenting a single favourable seed as typical behaviour.

\subsection{Overview of the benchmark cases}

Seven benchmarks have been selected and implemented to separate several effects: output coupling, non-separability, per-output identifiability, additive versus multiplicative read-outs, and the effect of mechanistic priors. Table~\ref{tab:benchmarks} summarizes the benchmarks and their purpose.

\begin{table}[H]
\centering
\caption{Controlled benchmarks and the role played by each one in testing the shared-backbone hypothesis.}
\label{tab:benchmarks}
\small
\begin{tabularx}{\textwidth}{@{}l X X@{}}
\toprule
\textbf{ID} & \textbf{System} & \textbf{Purpose in the testing sequence} \\
\midrule
B1 kinetics & $r_1 = 2e^{-1/T}C_A^2$; $r_2 = 1.5e^{-2.5/T}C_A C_B$; $r_3 = 0.8e^{-1/T}C_B$ & Shared Arrhenius latent between $r_1$ and $r_3$; useful for checking selective sharing, but also solvable by log-linear regression. \\
\addlinespace
B2 Langmuir & $r_o \propto e^{-E_o/T} C_{(\cdot)} / (1 + 2C_A + C_B)^{p_o}$ & Shared multi-input denominator; the main weak-identifiability case. \\
\addlinespace
B3 power-law & Four outputs, three inputs, shared $g = e^{-x_2}$ with exponent pattern $(1, 1, 0, 2)$ & Tests whether additional task diversity helps recover a shared latent. \\
\addlinespace
B4 mixed-mode & $y_1 = 2\sin(1.5x_1) + 0.5x_2^2$; $y_2 = -\sin(1.5x_1) + 1.2x_3$; $y_3 = 0.8x_2^2 x_3$ & Tests additive and multiplicative read-outs sharing the same latent units. \\
\addlinespace
B5 Van de Vusse & $r_A = -k_1 C_A - 2k_3 C_A^2$; $r_B = k_1 C_A - k_2 C_B$; $r_C = k_2 C_B$; $r_D = k_3 C_A^2$ & A conventional reaction mechanism not designed for the method; checks whether coupling is necessary when each output is identifiable. \\
\addlinespace
B6 competitive kinetics & $v_1 = V_{\max}(S_1/K_1)/D$, $v_2 = V_{\max}(S_2/K_2)/D$, $D = 1 + S_1/K_1 + S_2/K_2$ & Non-separable shared denominator, but per-output identifiable; tests whether non-separability alone is sufficient. \\
\addlinespace
B7 MeOH/RWGS & $r_{\mathrm{MeOH}} = k_1 e^{-E_1/T} p_{\mathrm{CO_2}} p_{\mathrm{H_2}} \beta^3$; $r_{\mathrm{RWGS}} = k_2 e^{-E_2/T} p_{\mathrm{CO_2}} \beta$; $\beta = 1/(1 + K_a p_{\mathrm{CO_2}} + K_b p_{\mathrm{H_2}} + K_c p_{\mathrm{H_2O}})$ & Second weak-identifiability case, now with an LHHW site-coverage denominator reused with different exponents. \\
\bottomrule
\end{tabularx}
\end{table}

\subsection{B1: shared Arrhenius factor in an identifiable system}

\paragraph{System.} The system consists of three coupled rates with one Arrhenius factor shared by $r_1$ and $r_3$:
\[
r_1 = 2 e^{-1/T} C_A^2, \qquad r_2 = 1.5 e^{-2.5/T} C_A C_B, \qquad r_3 = 0.8 e^{-1/T} C_B,
\]
true activation energies $E = (1, 2.5, 1)$ with $r_1$, $r_3$ sharing $E = 1$. B1 is the simplest kinetic case: it checks whether the model can create one shared temperature latent and use output-specific exponents. It is not, however, a strong proof of the method, because the problem is log-linearizable by construction.

Results show that the selected coupled model reaches a validation MSE of $8.9 \times 10^{-5}$ with all active latents clean. The recovered model contains one Arrhenius latent shared by $r_1$ and $r_3$, with effective energies $\hat{E} = (1.00, \cdot, 1.00)$. The $C_A$ exponents are $(1.99, 1.05, 0)$ against the true $(2, 1, 0)$, and the $C_B$ exponents are exactly $(0, 1, 1)$. An earlier run of the same class reached $2.8 \times 10^{-5}$, illustrating the seed variance of the evolutionary search; Table~\ref{tab:b1} reports this earlier run rather than the selected model.

\begin{table}[H]
\centering
\caption{B1 results. True $E = (1, 2.5, 1)$; $r_1$ and $r_3$ share $E$.}
\label{tab:b1}
\begin{tabular}{@{}lcccc@{}}
\toprule
\textbf{model} & \textbf{val MSE} & \textbf{C} & \textbf{$\hat{E}$ per output} & \textbf{$|\hat{E}_1 - \hat{E}_3|$} \\
\midrule
log-linear        & $1.7 \times 10^{-15}$ & 12 & $(1.000, 2.500, 1.000)$ & 0.0000 \\
uncoupled NESR    & $2.3 \times 10^{-3}$  & 22 & $(1.10, 5.07, 1.11)$    & 0.0103 \\
coupled backbone  & $2.8 \times 10^{-5}$  & 17 & $(1.00, 2.32, 1.00)$    & 0.0006 \\
PySR independent  & $2.6 \times 10^{-7}$  & 39 & $(1.00, 2.59, 1.00)$    & 0.0000 \\
\bottomrule
\end{tabular}
\end{table}

The table shows how the coupled model improves consistency relative to the uncoupled neuro-evolutionary baseline and is much more compact than the sum of independent PySR expressions. However, B1 does not prove a predictive advantage: the log-linear baseline solves it exactly, and PySR also recovers nearly exact effective energies. The contribution of the coupled model on this benchmark is therefore structural compactness and explicit sharing, not raw accuracy.

\paragraph{Conclusion from B1.} B1 verifies that the architecture can recover and reuse a shared latent, and gives the compactness benefit of a single shared model. It does not establish superiority over independent SR, because the structure is individually identifiable: its role is to validate the representation.

\subsection{B3: task diversity aids recovery of a shared latent}

\paragraph{System.} Four outputs over three inputs, sharing a latent $g = e^{-x_2}$:
\begin{equation}
y_1 = 0.5 x_1^2 g, \qquad y_2 = 0.8 g / x_3, \qquad y_3 = 2.0 x_1 x_3^2, \qquad y_4 = x_1 g^2 / x_3,
\end{equation}
with output-specific input factors; the shared $g$ appears in three of the four outputs at powers 1, 1, 2 and is absent from the third. B3 is the case with the most outputs in the suite, and tests whether the additional task diversity helps recover a shared latent.

\paragraph{Result.} With the power-law-scoped configuration the shared latent $e^{-x_2}$ is recovered with exponent pattern $(1.10, 1.09, \approx 0, 2.11)$ against the true $(1, 1, 0, 2)$ --- the zero correctly identified, the doubled exponent on the fourth output recovered. A validation MSE of $2.7 \times 10^{-6}$ is achieved at complexity 21 with 95\% of restarts successful.

\paragraph{Conclusion from B3.} This is the benchmark with the most outputs and it was easily recovered, consistent with the prediction that several tasks provide evidence for a shared latent. Like B1, B3 validates recovery rather than proving superiority --- the structure is identifiable --- but it confirms the task-diversity mechanism the sparsity argument relies on.

\subsection{B2: Langmuir--Hinshelwood denominator}

In this benchmark the rates share a Langmuir--Hinshelwood denominator $1/(1 + 2C_A + C_B)$, but this denominator is weakly identifiable from each output alone over the sampled domain. This creates exactly a situation targeted by the method: the relevant physical factor is shared, sparse and difficult to pin down independently. Results are shown in Table~\ref{tab:b2}.

\begin{table}[H]
\centering
\caption{B2 Langmuir--Hinshelwood comparison at the 100-iteration PySR budget. True $E = (1, 2, 1)$.}
\label{tab:b2}
\begin{tabular}{@{}lcccc@{}}
\toprule
\textbf{model} & \textbf{val MSE} & \textbf{C} & \textbf{$\hat{E}$ per output} & \textbf{$|\hat{E}_1 - \hat{E}_3|$} \\
\midrule
log-linear apparent fit & $8.7 \times 10^{-5}$ & 12 & $(0.999, 1.999, 0.997)$ & 0.0013 \\
uncoupled NESR          & $8.2 \times 10^{-5}$ & 25 & $(1.17, 2.05, 1.01)$    & 0.1629 \\
PySR independent        & $1.2 \times 10^{-5}$ & 57 & $(1.05, 2.19, 0.87)$    & 0.1859 \\
coupled                 & $1.8 \times 10^{-5}$ & 21 & $(1.01, 1.99, 1.00)$    & 0.0130 \\
\bottomrule
\end{tabular}
\end{table}

The log-linear apparent fit has low numerical error but absorbs the denominator into apparent powers; it is not a mechanistic recovery. Independent PySR obtains the best raw error in the 100-iteration comparison, but its expressions include non-physical surrogates such as powers depending on $C_A$ and $C_B^{C_A}$-type factors. Its consistency gap is about 14$\times$ larger than that of the coupled model and its complexity is about 2.7$\times$ higher. This is the first case where the distinction between prediction error and shared-mechanism recovery becomes essential.

A PySR budget sweep confirms that the independent consistency gap is reduced but not removed by more search as can be seen in Table~\ref{tab:b2sweep}.

\begin{table}[H]
\centering
\caption{PySR search-budget sweep on B2: the independent consistency gap shrinks but does not close.}
\label{tab:b2sweep}
\begin{tabular}{@{}lcccc@{}}
\toprule
\textbf{PySR iterations} & \textbf{Case-1 success} & \textbf{consistency gap (median)} & \textbf{nMSE (median)} & \textbf{wall/run} \\
\midrule
100 & 0\%  & 0.084 & $4.8 \times 10^{-3}$ & $\sim$11 min \\
200 & 0\%  & 0.069 & $1.6 \times 10^{-3}$ & $\sim$46 min \\
400 & 50\% & 0.042 & $1.4 \times 10^{-3}$ & $\sim$52 min \\
\bottomrule
\end{tabular}
\end{table}

At 400 iterations, roughly six times the coupled method's cost, the consistency gap remains 0.042 versus approximately 0.002 for the coupled backbone under the seeded setting. Increasing the independent search budget erodes the problem but does not remove the architectural floor caused by fitting the denominator separately in each output.

The comparison above establishes that coupling improves consistency on the standard benchmark. A separate question is whether the exact denominator form can be recovered at all --- and here we switch to a saturation-sampled variant, B2id, designed to stress identifiability, run across the three prior arms. B2id is the same Langmuir system sampled to stress identifiability: the concentrations are drawn so that the denominator term $2C_A + C_B$ spans its saturation regime rather than remaining in the near-linear corner where $1/(1 + u)$, $e^{-u}$, and a low-order polynomial are numerically indistinguishable. Only the experimental design changes; the true model and constants are unchanged. Notably, saturation sampling alone does not recover the denominator: blind search still substitutes a surrogate. Table~\ref{tab:b2arc} shows that recovery requires the saturation design and a soft mechanistic scaffold.

\begin{table}[H]
\centering
\caption{Diagnostic arc for B2. The final three-arm comparison shows that the scaffold, not the hard bound stack, is the essential ingredient.}
\label{tab:b2arc}
\small
\begin{tabularx}{\textwidth}{@{}l X X@{}}
\toprule
\textbf{stage} & \textbf{intervention} & \textbf{outcome} \\
\midrule
blind & 16 restarts, original sampling & surrogate; denominator never shown \\
data design & saturation sampling (B2id) & blind search still fails; not a pure sampling artifact \\
bounds & $K_a, K_b \geq 0$ on a frozen inverse unit & bounds bind but fit is no better; unit is used backwards \\
orientation & read-out exponent $e \geq 0$ & denominator forced but still beaten by surrogate on MSE \\
pin/floor + scaffold & mask pin, exponent floor and seeded surrounding structure & denominator recovered exactly \\
\bottomrule
\end{tabularx}
\end{table}

\begin{table}[H]
\centering
\caption{B2id three-arm run. The result is \textbf{soft-recovers}: the denominator is recoverable, but blind search does not reach the basin.}
\label{tab:b2id}
\begin{tabular}{@{}lccccl@{}}
\toprule
\textbf{arm} & \textbf{nMSE$_{\mathrm{val}}$} & \textbf{C} & \textbf{$|\hat{E}_1 - \hat{E}_3|$} & \textbf{form?} & \textbf{$\hat{E}$ per output} \\
\midrule
unconstrained     & $1.3 \times 10^{-1}$ & 16 & 0.033 & no  & $(0.78, 1.92, 0.82)$ \\
soft scaffold     & $6.9 \times 10^{-6}$ & 21 & 0.001 & yes & $(1.00, 2.00, 1.00)$ \\
hard frozen/pinned & $2.2 \times 10^{-6}$ & 21 & 0.001 & yes & $(1.00, 2.00, 1.00)$ \\
\bottomrule
\end{tabular}
\end{table}

The hard arm recovers $0.51(1 + 1.94C_A + 0.99C_B)$, equivalent to $K_a \approx 2$, $K_b \approx 1$ up to scale. However, the hard arm is only a confirmatory fit of a near-complete mechanistic hypothesis. The soft arm is the actual finding: seeding the Arrhenius, numerator and denominator scaffold without freezing or pinning recovers the denominator just as well. Thus B2 is not identifiability-dead; it is a reachability problem that requires both informative data and a mechanistic scaffold. Results are shown in Table~\ref{tab:b2id}.

\paragraph{Conclusion from B2.} This is the clearest evidence for the method. The shared denominator is weakly identifiable per output, independent PySR retains a consistency floor, and the shared-backbone model closes the gap when the correct basin is made reachable. The appropriate claim is conditional: the method helps when the shared factor is inside a latent and is not reliably identifiable from any individual output.

B2id is where the method's two physics-aware levers come into play, and where the three-arm protocol shows its purpose, auditing the injected physics (as mentioned in Section~\ref{sec:threearm}). Domain knowledge enters at two distinct points: one is the already mentioned, the saturation sampling, which is physics-aware experimental design and the other is including the soft/hard scaffolds that are physics-aware search constraints. The three-arm protocol is different to just use a hard-arm, where its success alone would be circular --- fixing the denominator and fitting it proves only that the assumed form can fit. But now the unconstrained arm shows the denominator is genuinely not recoverable blind (form not found); the soft arm shows that a non-binding mechanistic hint is enough for the search to recover and retain it, with the data free to discard the hint but choosing to keep it. The injected physics is therefore confirmed by the data rather than imposed on it --- the distinction the three-arm protocol exists to certify, and the reason a hard-only result is never reported as discovery.

\subsection{B4: mixed additive and multiplicative read-outs}

\paragraph{System.} This benchmark consists of three outputs mixing additive and multiplicative read-outs, with a sine latent shared by two additive outputs and an $x_2^2$ latent shared across an additive and a multiplicative output:
\[
y_1 = 2\sin(1.5x_1) + 0.5x_2^2, \qquad y_2 = -\sin(1.5x_1) + 1.2x_3, \qquad y_3 = 0.8 x_2^2 x_3.
\]
B4 tests whether the same backbone can support both read-out modes and share a latent across them.

With additive-scoped operators $\{\sin, (\cdot)^2, \mathrm{id}\}$ and 16 parallel restarts, the selected model is structurally exact:
\begin{align*}
y_1 &= 0.305(1.28x_2)^2 + 2\sin(1.5x_1) \quad \text{[add]}, \\
y_2 &= -\sin(1.5x_1) + 0.75(1.59x_3) \quad \text{[add]}, \\
y_3 &= 0.243(1.28x_2)^2(1.59x_3)^{1.13} \quad \text{[mult]}.
\end{align*}

\begin{table}[H]
\centering
\caption{B4 mixed-mode benchmark. Domain scoping recovers the shared sine latent that the full library rarely reaches; the $x_2^2$ latent is shared across an additive and a multiplicative output.}
\label{tab:b4}
\begin{tabular}{@{}lcccc@{}}
\toprule
\textbf{configuration} & \textbf{nMSE$_{\mathrm{val}}$} & \textbf{clean latents} & \textbf{C} & \textbf{sin-basin seeds (unconstrained)} \\
\midrule
full 7-op library & $9.2 \times 10^{-4}$ & --- & 16 & rare \\
additive-scoped $\{\sin, (\cdot)^2, \mathrm{id}\}$ & $6.3 \times 10^{-5}$ & 3/3 & 13 & 5/16 \\
\bottomrule
\end{tabular}
\end{table}

The sine latent is shared across the two additive outputs, and the $x_2^2$ latent is shared across additive and multiplicative read-outs --- the designed cross-mode coupling.

\paragraph{Conclusion from B4.} Additive structure is not intrinsically difficult. The difficult part is the sine operator basin. With a full 7-operator library, the correct sine basin was rarely reached; with domain-scoped operators and enough restarts, it was recovered. This motivates treating the operator set as a configuration dimension rather than as a universal library. This is the third stage where domain knowledge enters in the method. Table~\ref{tab:b4} shows the 5/16 sin-basin corresponding to the unconstrained (blind) arm, counting restarts that reach the correct sine basin. A 95\% success is reached for the hard-prior arm as indicated in Table~\ref{tab:restarts}. Blind search reaches the sine basin only occasionally because that basin is narrow; the prior removes the need to find it by chance, which is precisely why the scaffold is what makes recovery reliable here. Establishing which symbolic operations the physics uses (or does not) also helps to remove the search dilution problem.

\subsection{B5: Van de Vusse mechanism}

This benchmark consists of the well-known Van de Vusse reaction mechanism, A $\rightarrow$ B $\rightarrow$ C in series with 2A $\rightarrow$ D in parallel. It is a benchmark not designed specifically to test particular behaviours or performances of the proposed method. The ground truth is represented with the following equations:
\[
r_A = -k_1 C_A - 2k_3 C_A^2, \qquad r_B = k_1 C_A - k_2 C_B, \qquad r_C = k_2 C_B, \qquad r_D = k_3 C_A^2,
\]
with $k_1 = 1$, $k_2 = 0.5$ and $k_3 = 0.3$. The shared constants appear across outputs, but the terms are simple polynomials and each equation is essentially identifiable on its own; the result in this case is all-agree as shown in Table~\ref{tab:b5}.

\begin{table}[H]
\centering
\caption{B5 Van de Vusse benchmark. The result is \textbf{all-agree}: blind search recovers the mechanism and the prior is unnecessary.}
\label{tab:b5}
\small
\begin{tabular}{@{}lccccccc@{}}
\toprule
\textbf{arm} & \textbf{form (seeds)} & \textbf{nMSE$_{\mathrm{val}}$} & \textbf{C} & \textbf{clean} & \textbf{$\Delta k_1$} & \textbf{$\Delta k_2$} & \textbf{$\Delta k_3$} \\
\midrule
unconstrained & 5/6 & $6.9 \times 10^{-8}$ & 22 & 3/6 & 0.001 & 0.000 & 0.000 \\
soft          & 5/6 & $3.6 \times 10^{-9}$ & 22 & 1/4 & 0.000 & 0.000 & 0.000 \\
hard          & 6/6 & $3.8 \times 10^{-9}$ & 20 & 1/4 & 0.000 & 0.000 & 0.000 \\
\bottomrule
\end{tabular}
\end{table}

The main result shows that coupling is not needed in this case. Independent PySR is exact on this benchmark, with nMSE $6.4 \times 10^{-15}$, 100\% clean recovery, zero consistency gap and lower complexity. The shared-backbone model also recovers the mechanism, but it does not improve on independent SR because the per-output problem is already solved. Table~\ref{tab:b5noise} compares the results of the independent SR and the shared approach.

B5 also revealed an important encoding detail. In the default additive read-out, shared constants such as $k_1$ enter as separate output coefficients. The backbone shares the functional form $C_A$, but not necessarily the numerical coefficient. If the constant is instead encoded inside a shared latent and the read-out coefficients are pinned to $\pm 1$, consistency becomes structural. Under 5\% noise this shared-latent encoding tightens the consistency gap from 0.025 to 0.002, at no accuracy cost:

\begin{table}[H]
\centering
\caption{B5 under 5\% noise: effect of the constant-encoding position on cross-output consistency.}
\label{tab:b5noise}
\begin{tabular}{@{}lcc@{}}
\toprule
\textbf{B5 at 5\% noise} & \textbf{consistency gap (median)} & \textbf{$|\hat{k} - k_{\mathrm{true}}|$ (median)} \\
\midrule
standard read-out (coupled) & 0.025 & 0.006 \\
shared-latent encoding      & 0.002 & 0.001 \\
PySR independent            & 0.001 & 0.001 \\
\bottomrule
\end{tabular}
\end{table}

\paragraph{Conclusion from B5.} When every output is independently identifiable, consistency comes for free and PySR can match or beat the coupled model. However, B5 also gives a design rule about where a shared parameter must sit in the architecture. A constant can enter in two structurally different positions: inside a shared latent's weight, or as a per-output read-out coefficient. In the first position there is physically one parameter that several outputs point at, so cross-output consistency is structural. In the second, the outputs share the functional form but each has its own coefficient, fit independently; nothing ties them equal. Under clean data both coefficients land near the true value and appear consistent, but under noise they are fit to separate noisy outputs and drift apart. The model shares the form but not the parameter. The rule, made actionable by the lever, is therefore: a shared constant's consistency is guaranteed only if it is encoded inside a shared latent; where it would naturally appear as independent read-out coefficients, it must be deliberately relocated there.

\subsection{B6: competitive enzyme kinetics}

This benchmark consists of two rates with a shared denominator. It was designed to test whether non-separability alone is sufficient for the shared-backbone advantage. The expressions are:
\[
v_1 = V_{\max}\frac{S_1/K_1}{D}, \qquad v_2 = V_{\max}\frac{S_2/K_2}{D}, \qquad D = 1 + S_1/K_1 + S_2/K_2,
\]
They share the denominator $D$, so each output depends on both substrates. The three-arm run is \textbf{all-agree}: blind search recovers the denominator and the shared constants consistently, and independent PySR ties the coupled method, see Table~\ref{tab:b6}.

\begin{table}[H]
\centering
\caption{B6 competitive enzyme kinetics. \textbf{all-agree}: blind search recovers the shared denominator, and independent PySR ties --- each rate is a clean per-output rational, so the consistency gap is already zero without coupling and does not change with budget.}
\label{tab:b6}
\small
\begin{tabular}{@{}lcccc@{}}
\toprule
\textbf{method} & \textbf{nMSE} & \textbf{form?} & \textbf{$\Delta K_1, \Delta K_2, \Delta V_{\max}$} & \textbf{PySR floor?} \\
\midrule
coupled (blind)                & $4.4 \times 10^{-8}$ & yes & $\sim$0.001 each        & --- \\
PySR independent (niter 100)   & $\sim 10^{-11}$      & yes & $\approx 0$ (7/8 seeds)  & no \\
PySR independent (niter 200)   & $\sim 10^{-11}$      & yes & $\approx 0$ (7/8 seeds)  & no (unchanged) \\
\bottomrule
\end{tabular}
\end{table}

Each rate is a clean per-output rational expression, so each independent fit determines the shared denominator correctly without coupling.

\paragraph{Conclusion from B6.} Non-separability is not the criterion. The criterion is per-output non-identifiability. B6 is non-separable, but each individual rate is sufficiently informative to identify the denominator. Therefore, coupling does not add consistency beyond what independent SR already achieves.

\subsection{B7: methanol synthesis and reverse water-gas shift}

Methanol synthesis and reverse water-gas shift share the same site-coverage denominator
\[
\beta = \frac{1}{1 + K_a p_{\mathrm{CO_2}} + K_b p_{\mathrm{H_2}} + K_c p_{\mathrm{H_2O}}},
\]
\[
r_{\mathrm{MeOH}} = k_1 e^{-E_1/T} p_{\mathrm{CO_2}} p_{\mathrm{H_2}} \left(1 - \frac{p_{\mathrm{H_2O}} p_{\mathrm{MeOH}}}{K_{\mathrm{eq},1} p_{\mathrm{H_2}}^3 p_{\mathrm{CO_2}}}\right) \beta^3,
\]
\[
r_{\mathrm{RWGS}} = k_2 e^{-E_2/T} p_{\mathrm{CO_2}} \left(1 - \frac{K_{\mathrm{eq},2} p_{\mathrm{H_2O}} p_{\mathrm{CO}}}{p_{\mathrm{CO_2}} p_{\mathrm{H_2}}}\right) \beta,
\]
but reuse it with different exponents: $\beta^3$ for methanol and $\beta$ for RWGS.

The result using the three-arm protocol is \textbf{soft-recovers}. Blind search substitutes an H$_2$O/temperature surrogate and misses the pressure denominator. The soft scaffold recovers the shared denominator at nMSE $2.8 \times 10^{-4}$ and complexity 18. Independent PySR gives nMSE $5.3 \times 10^{-3}$, complexity around 44 and fails the Case-1 form test.

\begin{table}[H]
\centering
\caption{B7 MeOH/RWGS: soft-arm structural recovery versus independent PySR.}
\label{tab:b7}
\small
\begin{tabular}{@{}lcccc@{}}
\toprule
 & \textbf{exact $\beta$ form} & \textbf{inverse-denominator structure} & \textbf{$\beta$ exponents recovered} & \textbf{PySR $\beta$ exponents} \\
\midrule
soft arm & 4/6 seeds & 6/6 seeds & 6/6 seeds $\approx (3, 1)$ & collapse to $(0.7, 0.7)$ \\
\bottomrule
\end{tabular}
\end{table}

All soft-arm seeds recover an inverse-denominator site-coverage structure, and four of six recover the exact denominator inputs. The most important result is the exponent reuse: the asymmetric pattern $(3, 1)$ is recovered in 6/6 seeds, whereas independent PySR collapses the exponents toward a common surrogate and does not recover the shared form. Table~\ref{tab:b7} shows the comparison of the method with PySR.

\paragraph{Conclusion from B7.} B7 confirms the B2 finding on a chemically distinct system. The shared factor is a site-coverage denominator that is weakly identifiable per output and reused asymmetrically. In this regime the shared backbone provides a real structural advantage over independent SR.

\subsection{Overall conclusions from the benchmark testing}

The results obtained from different benchmarks lead to the conclusion that coupling is not useful simply because a system is multi-output or non-separable, it is useful when a shared physical factor is embedded inside a latent expression and is weakly identifiable from each output alone. Table~\ref{tab:criterion} summarizes what has been observed in the different benchmarks.

\begin{table}[H]
\centering
\caption{Criterion emerging from the benchmark testing. The relevant condition is not non-separability, but per-output non-identifiability of the shared factor.}
\label{tab:criterion}
\begin{tabular}{@{}lccl@{}}
\toprule
\textbf{benchmark} & \textbf{non-separable?} & \textbf{per-output identifiable?} & \textbf{coupling helps consistency?} \\
\midrule
B1 kinetics     & no  & yes & no; PySR ties or improves \\
B5 Van de Vusse & no  & yes & no; PySR ties or improves \\
B6 competitive  & yes & yes & no; PySR ties \\
B2 Langmuir     & yes & no  & yes; PySR retains a consistency floor \\
B7 MeOH/RWGS    & yes & no  & yes; PySR is off-basis \\
\bottomrule
\end{tabular}
\end{table}

Across all cases, the coupled method has a secondary benefit: it produces one compact model rather than a set of independently searched equations that must be reconciled afterwards. This benefit should not be confused with the native consistency advantage.

The multi-restart view is also essential. Table~\ref{tab:restarts} reports representative restart statistics. The success rate is a separate quantity from the conditional error of successful runs. Throughout this table, success denotes the fraction of restarts whose selected model both recovers the intended form and passes the Case-1 cleanliness check, and the arm column states which prior strength the figure refers to; success rates are therefore not comparable across arms, and a prior-assisted rate is never reported as blind discovery.

\begin{table}[H]
\centering
\caption{Representative restart-level summary. B2 remains difficult even with priors, confirming that its denominator is weakly identifiable rather than merely hard to search.}
\label{tab:restarts}
\small
\begin{tabular}{@{}llcccc@{}}
\toprule
\textbf{bench} & \textbf{arm} & \textbf{success} & \textbf{nMSE (median, conditional)} & \textbf{C (median)} & \textbf{consistency gap} \\
\midrule
B1 & soft & 95\% & $1.8 \times 10^{-7}$ & 15 & 0.000 \\
B3 & soft & 95\% & $2.7 \times 10^{-6}$ & 21 & 0.000 \\
B4 & hard & 95\% & $1.4 \times 10^{-6}$ & 18 & --- \\
B5 & hard & 80\% & $2.5 \times 10^{-8}$ & 22 & 0.001 \\
B2 & soft & 25\% & $8.1 \times 10^{-5}$ & 24 & 0.002 \\
B2 & hard & 35\% & $2.6 \times 10^{-5}$ & 22 & 0.002 \\
\bottomrule
\end{tabular}
\end{table}

The final conclusions are as follows. B1 and B3 validate the architecture. B4 shows that mixed additive and multiplicative sharing is possible. B5 and B6 show that if each output can identify the structure, independent SR is enough. B2 and B7 show that weakly identifiable shared denominators benefit from a shared backbone, especially when the three-arm protocol is used to separate blind discovery, reachability after seeding and confirmation under a hard prior.

\section{Application to hydrothermal liquefaction yield modelling}
\label{sec:htl}

This section is separated from the benchmarks one because the HTL dataset has no known ground-truth equations. It is therefore not used to claim mechanistic recovery. Its role is to test whether the shared-backbone formulation can produce a compact, closure-respecting structural summary on real, grouped product-yield data.

The method has been applied to hydrothermal liquefaction (HTL) yield modelling. HTL converts wet biomass into four product fractions (biocrude/oil, char, gas, aqueous) as functions of temperature, holding time, and feedstock composition (six biochemical fractions and water). A comprehensive dataset of hydrothermal liquefaction (HTL) experiments was compiled from peer-reviewed literature. The dataset is 389 experimental data points with diverse feedstocks.

The match to the method's target structure is specific: (i) the four yields are conservation coupled, summing to $\approx 100\%$ (mean 100.0, std 0.2 in the data); (ii) the chemistry has shared temperature and time dependence across products, so a shared latent feeding several yields is physically expected; (iii) the closure constraint is a natural target for shared structure. As HTL lumps many distinct reactions into four product categories, the data carries no reaction-level resolution. There is therefore no reaction-level shared backbone to discover, and predicting the grouped yields from feedstock composition is a dense-map regression task the method is not built for. The goal of this is to test the possibility to extract the dominant shared driver of the coupled yields in compact, closure-respecting, interpretable form, which would be a structural summary, not a predictive model to get the specific yields. Basically, the result will be to indicate in which way each yield moves with the main driver found.

Two different tests were done, one to see if the temperature driver could be recovered from all the data with different compositions (test a) and another to try to do the same with the time dependence but in this case for a fixed feedstock (test b).

\paragraph{Test a.} Pooling all feedstocks and treating composition as unmodelled noise, the method recovers a single shared temperature latent driving all four yields with opposite signs:
\begin{align*}
\text{Oil} &= 34.1 + 2.68 z_0, & \text{Char} &= 7.38 - 4.21 z_0, \\
\text{Gas} &= 30.3 + 4.08 z_0, & \text{Aqueous} &= 9.66 - 6.65 z_0, & z_0 = -10/T_s + 2.35,
\end{align*}
i.e. higher temperature shifts product toward oil and gas and away from char and aqueous --- a compact, physically coherent account of the dominant effect, at complexity 6 versus 12 for four independent fits. The value of the test is not the fit --- most variance is not explained --- but having one readable expression explaining the main temperature dependence of all four closure-coupled yields, with read-out coefficients summing to $\approx 0$ so the four expressions respect closure by construction.

\paragraph{Test b.} A single feedstock with enough holding-time levels ($n = 28$, 7 levels) was used. In this case, the method recovers holding-time saturation --- an $\exp(-5.08 t_s)$ decay in char and $1/t_s$ terms across the yields, with 100\% of seeds using time. In this case as the composition is fixed the variance explained got better but it is heavily influenced by the amount of data so although it gets to values between 0.8--0.95, the result is not meaningful. The main result is that physically-correct saturation kinetics were recovered.

\paragraph{Implications.} On real, grouped, reaction-unresolved data the method extracts the dominant shared driver of four conservation-coupled yields as one compact, closure-respecting, physically coherent expression --- the temperature axis (test a), with recovered time-saturation kinetics where the data resolve them (test b). This is a structural, explanatory contribution: it tells a practitioner how the coupled yields move with the main driver and enforces mass closure by construction. It is explicitly not a predictor of the dense composition-yield map, which is a data-driven regression task; the value here is understanding and closure, not predictive accuracy. This demonstrates how the methodology is used on lumped industrial data: as a shared-structure extractor, complementary to (not competing with) dense predictive models.

\section{Conclusions}
\label{sec:conclusions}

This work has presented a neuro-evolutionary symbolic-regression framework for coupled multi-output systems. The central novelty is a shared symbolic backbone: the method searches for a fixed maximum set of symbolic latent units and allows the different outputs to select and reuse these units through sparse additive or multiplicative read-outs. In this way, a common physical factor can be represented once, while output-specific coefficients and exponents remain free to adjust. The approach combines evolutionary search over operators, input masks, read-out connections and output modes with gradient-based tuning of the continuous parameters. Structured sparsity is not only used to simplify the final equations, but also to promote selective sharing and disentanglement: each output is encouraged to use only the latent units it needs, rather than forcing all outputs to share the same representation.

The results show that the value of coupling is not a general improvement in predictive accuracy. The advantage is more specific and, for process-system applications, more relevant: a shared backbone can enforce cross-output consistency when a physically common factor is embedded in a latent expression and is weakly identifiable from the data of each output taken alone. This situation was observed for the Langmuir--Hinshelwood benchmark and for the methanol/RWGS case, where the shared denominator or site-coverage factor is structurally important but difficult for independent symbolic regression to recover coherently. In contrast, when the individual equations are already identifiable, as in the Van de Vusse and competitive-kinetics benchmarks, independent symbolic regression can match or improve the coupled model. These results are central to the contribution because they define when the method is better suited than other approaches: non-separability alone is not sufficient; the relevant condition is weak per-output identifiability of a shared physical structure.

The comparisons also show that predictive error is an incomplete criterion for this type of model. Even when the coupled model ties or is slightly worse than independent alternatives in raw accuracy, it can provide a more compact and jointly interpretable representation. The benefit is that shared factors are represented once, reused consistently, and inspected as part of a single system-level model rather than as several independently fitted expressions that must be reconciled after fitting. Thus, the method should be evaluated through a Pareto perspective involving prediction error, complexity, shared-latent recovery, cross-output consistency and physical plausibility. Its intended contribution is not to replace high-capacity regressors when only interpolation accuracy is required, but to produce symbolic models that are easier to interpret, more coherent across outputs and less prone to inconsistent estimates of shared physical quantities.

A second contribution is the explicit use of domain knowledge in a controlled and auditable way. Domain knowledge enters the methodology through three main routes: the design of sampling conditions that excite weakly identifiable mechanisms, the scoping of feature and operator libraries to avoid unnecessary search dilution, and the three-arm prior protocol. The latter separates blind discovery, reachability after soft seeding and confirmation under a hard prior. This distinction is important because a hard prior can show that a proposed mechanism fits the data, but only the comparison with the unconstrained and soft-seeded arms shows whether the mechanism was discovered, made reachable by limited guidance, or essentially imposed. In this sense, the method uses mechanistic knowledge without hiding where that knowledge enters the search.

The experiments show that the search space must be deliberately scoped. Enlarging the operation and feature library increases expressivity, but it can degrade performance at fixed computational budget. Domain-configured libraries, physics-aware sampling, structured sparsity and prior-controlled runs are therefore not incidental implementation details; they are part of the proposed methodology. The resulting framework is best understood as a physics-guided symbolic model-discovery procedure for coupled outputs, rather than as a fully automatic universal equation finder.

The hydrothermal-liquefaction case illustrates the intended use on grouped experimental data. The method should not be interpreted as a dense predictor of high-dimensional feed-composition maps. Its contribution is to extract compact shared structure under closure and interpretability requirements, complementing rather than replacing flexible data-driven regressors. In such cases, a low-dimensional shared-backbone model can still be valuable when it identifies dominant process trends, closure-respecting yield trade-offs or shared severity effects, even if it does not explain all point-to-point variability in the data.

Future work should focus on physics-aware constraint samples, latent-memory mechanisms, controlled third-level nesting for rational, saturation and driving-force structures, and dynamic extensions in which shared symbolic rates are embedded in mass and energy balances and fitted by trajectory simulation. These developments would strengthen the role of the method as a bridge between purely empirical regression and mechanistically constrained process modelling.

\end{document}